\documentclass{alaya}
\usepackage[sc]{mathpazo}
\usepackage{amsmath,amsfonts,bm}

\def\eqref#1{equation~\ref{#1}}

\def\1{\bm{1}}

\DeclareMathAlphabet{\mathsfit}{\encodingdefault}{\sfdefault}{m}{sl}
\SetMathAlphabet{\mathsfit}{bold}{\encodingdefault}{\sfdefault}{bx}{n}

\usepackage[round]{natbib}
\usepackage{url}
\usepackage{tabularx}
\usepackage{siunitx}
\newcommand{\snd}[2]{\multicolumn{1}{c}{\setbox0\hbox{#1}\makebox[\wd0][r]{\underline{#2}}}}
\usepackage{tikz}
\usetikzlibrary{arrows.meta,positioning,fit}
\usepackage{environ}

\graphicspath{{figures/}}


\newcommand{\modelname}{PAWA}

\title{\modelname{}: Panoramic Action-conditioned World Modeling for Multiple Agents}

\author[1,2,*]{Siqi Yang}
\author[2,*]{Yimu Wang}
\author[1,2,3]{Ziqi Cai}
\author[2]{Chengxuan Zhu}
\author[2]{Shuchen Weng}
\author[3]{Erwin Wu}
\author[2]{Zhaofei Yu}
\author[1,\ddagger]{Kaipeng Zhang}
\author[2,\dagger]{Boxin Shi}

\affiliation[1]{Alaya Lab}
\affiliation[2]{Peking University}
\affiliation[3]{Institute of Science Tokyo}

\contribution[*]{Equal contribution}
\contribution[\dagger]{Corresponding author}
\contribution[\ddagger]{Project leader}

\newcommand{\pawacodelink}{https://github.com/yousiki/pawa-pipeline}
\newcommand{\pawadatalink}{https://huggingface.co/datasets/YOUSIKI/pawa-dataset}
\newcommand{\pawareleaseabstract}{The dataset and collection pipeline are open source to
support further research on multiplayer world models.}
\newcommand{\pawareleaseintro}{Both are open source on \href{\pawadatalink}{Hugging Face} and
\href{\pawacodelink}{GitHub}.}

\let\alayaabstract\abstract
\RenewEnviron{abstract}{\global\let\pawaabstract\BODY}
\providecommand{\pawareleaseabstract}{We will release the dataset and collection pipeline to
support further research on multiplayer world models.}
\begin{abstract}
A multiplayer world model must predict how the independent actions of several players change
a shared environment, including changes outside a player's current field of view.
A point-of-view (POV) video captures such changes only after they enter its field of view, whereas
panoramic video records the surroundings in every viewing direction and has served as an
intermediate representation of the world state in recent world models. Learning this task
therefore requires synchronized POV and panoramic videos of interacting players together with
their actions, yet no existing dataset provides this combination in Minecraft, an
established testbed in which players act independently in a shared, modifiable 3D world. We
present \modelname{}, a dataset and framework for action-conditioned panoramic multiplayer world
modeling. The dataset pairs each player's POV video with world-stabilized panoramic video and
aligned actions. Its distributed pipeline decouples CPU gameplay capture from GPU rendering,
scales beyond 64 simultaneous players, and re-renders each recorded session under new observation
configurations without replaying the gameplay. On this dataset, we develop a dual-tower model
that jointly predicts POV and panoramic video and couples the two streams through cross-attention
that aligns tokens observing the same direction in a shared world frame. We further introduce a
quadratic-phase player encoding that provides many more player codes than existing rotary
codes at the same embedding dimension. Experiments show that \modelname{} outperforms
state-of-the-art multiplayer world models in POV prediction quality while also producing
high-fidelity panoramic predictions. \pawareleaseabstract{}
\end{abstract}

\alayaabstract{\pawaabstract\par\vspace{0.3cm}}

\project{\url{https://pawa.siki.moe}}
\code{\expandafter\url\expandafter{\pawacodelink}}
\metadata[Dataset]{\expandafter\url\expandafter{\pawadatalink}}
\date{\today}

\begin{document}

\thispagestyle{firstheader}
\maketitle
\pagestyle{fancy}

\providecommand{\pawareleaseintro}{We will release both to support future multiplayer world models.}
\suppressfloats[t]
\begin{figure}[t]
\centering
\includegraphics[width=\linewidth]{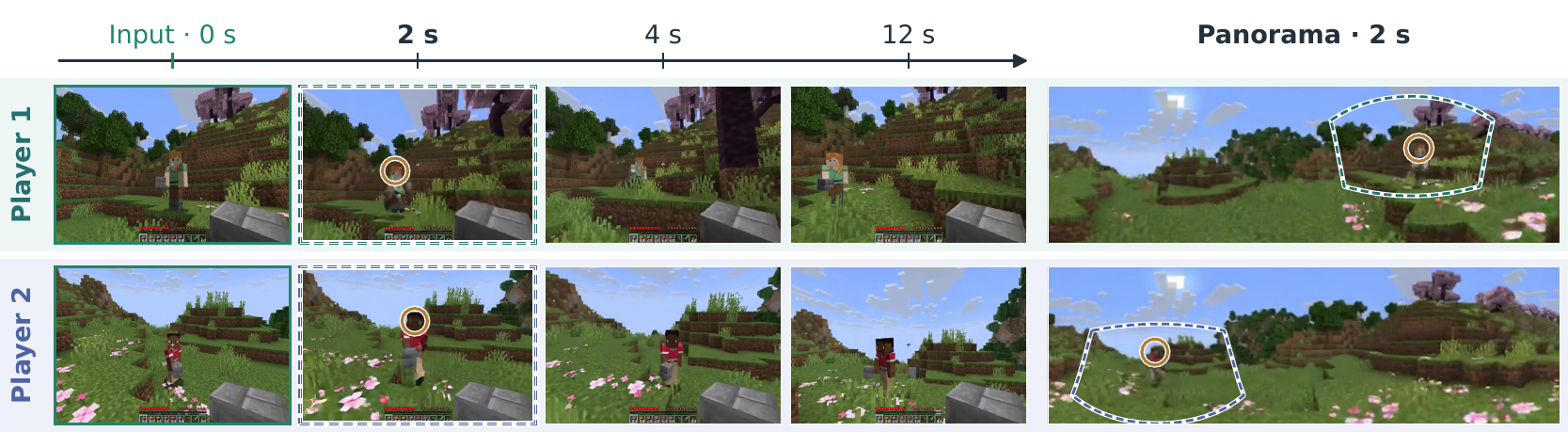}
\caption{\textbf{Panoramic multiplayer world modeling with \modelname{}.}
From each player's initial POV and
panoramic observations and the actions of both players, \modelname{} jointly forecasts every
player's POV video and world-stabilized panoramic video. Apart from the green-bordered inputs,
all frames are forecasts of the distilled model. The dashed outline projects the dashed POV
frame at 2\,s onto the panorama at the same time. Circles mark the co-player, which appears in the matching direction in the POV and the panorama. Panoramas are cropped to $\pm55^\circ$ latitude for display.}
\label{fig:teaser}
\end{figure}

\section{Introduction}
\label{sec:intro}

In a multiplayer world, several players act independently on a shared environment, and each
action can change what every other player sees. A world model for such a world must predict how
these combined actions unfold in each player's observations. Many of the consequences arise
outside a player's field of view. A teammate places a block behind the player, or an opponent
approaches from the side, and the change becomes visible only once the camera turns toward it.
A point-of-view (POV) video therefore records the shared world only in part. Panoramic video
records the surroundings in every direction at once, and recent world models have adopted it as
a representation of the surrounding scene and as spatial
memory~\citep{jiang2026panoworld,li2026panoworldmemory}. \emph{Pairing each player's POV video
with a panorama exposes the surroundings of every player to the model, including the regions
that no local camera currently observes.}

Minecraft offers a natural setting, since players act independently in a
shared 3D world that every action can modify. It is an established testbed for
action-conditioned video world
models~\citep{guo2025mineworld,yu2025gamefactory,zhang2025matrixgame,he2025matrixgame2}, and
recent multiplayer models synchronize the POV videos of several interacting
players~\citep{savva2026solaris,liu2026gammaworld,wu2026multiworld}. These multiplayer models
learn from perspective observations alone. Other multi-agent world models broaden visual
coverage by stitching directional views into panoramas~\citep{zhu2026shareverse} or by
projecting panoramic recordings into perspective training views~\citep{sun2026prismaworld}, yet
their generated streams remain perspective videos. \emph{Predicting full-sphere panoramic
video jointly with each player's POV under the actions of interacting players remains
unexplored.}

Modeling POV and panoramic video together for many players raises three challenges.
\textbf{\textit{(i) Paired multiplayer data.}} Existing multiplayer data are recorded by screen
capture from live game clients, which yields only the displayed perspective view. Every player
also requires its own graphics client, and the recording cost therefore grows with the
population. No dataset pairs POV and panoramic video with the actions of interacting players.
\textbf{\textit{(ii) Correspondence across projections.}} The same scene direction occupies
different image coordinates in a rectilinear POV frame and an equirectangular panorama. Because
the POV camera turns with the player's head while the panorama stays fixed to the world, this
correspondence changes at every frame.
\textbf{\textit{(iii) Player identity at scale.}} A model that shares parameters across players
must keep each observation stream bound to its own actions. Existing rotary player codes admit
only $m+1$ equidistant identities in $m$ phase planes~\citep{liu2026gammaworld}, and the
identity pool therefore grows only linearly with the embedding dimension.

In this paper, we present \modelname{} (\textbf{P}anoramic \textbf{A}ction-conditioned
\textbf{W}orld modeling for multiple \textbf{A}gents), a dataset and framework for action-conditioned panoramic
multiplayer world modeling in Minecraft.
\textbf{\textit{(i) Scalable, programmable collection.}} \modelname{} records gameplay on CPUs
without graphics clients, simulating sessions with more than 64 simultaneous players, and renders
the saved replays offline on GPUs. Each replay yields a player's POV video, a world-stabilized
equirectangular panorama and aligned actions, and can be rendered again with relative depth or
along custom camera trajectories without repeating gameplay. These observations serve
multiplayer world modeling well beyond panoramic prediction. Coding agents compose interaction
scripts from natural-language descriptions, and behavioral checks together with vision-language
review retain the scripts whose intended interactions occur (Figure~\ref{fig:collection-overview}).
\textbf{\textit{(ii) Dual towers with ray-aligned cross-attention.}} A POV tower and a panorama tower jointly
predict the two videos and are repeatedly coupled through bidirectional ray-aligned
cross-attention. Each token carries its viewing ray in a shared world frame as a rotary phase, and tokens
that observe the same direction thus have zero relative phase regardless of their image coordinates
(Figure~\ref{fig:model-overview}). The panorama thus serves as both a prediction target and a
source of context for the local view. In the resulting forecasts, a co-player seen in a player's
POV appears in the matching direction of the same player's panorama (Figure~\ref{fig:teaser}).
\textbf{\textit{(iii) Quadratic-phase player encoding.}} Building on classical quadratic-phase
codes~\citep{wootters1989optimal}, \modelname{} assigns each player a rotary code from a pool of
$m^2$ identities in the same $m$ phase planes. The identity pool thus grows quadratically rather than linearly with the embedding
dimension.

Our contributions are threefold. \textbf{(i)} We construct a multiplayer Minecraft dataset of
synchronized POV and world-stabilized panoramic videos with aligned actions, whose distributed
pipeline scales beyond 64 simultaneous players and re-renders replays with relative depth or
custom camera trajectories. \pawareleaseintro{}
\textbf{(ii)} We propose a dual-tower world model that jointly predicts POV and panoramic video,
whose ray-aligned cross-attention encodes world-frame viewing rays as rotary phases and thereby
assigns identical phases to matching viewing directions in the two projections.
\textbf{(iii)} We introduce a quadratic-phase player encoding that enlarges the identity pool from
$m+1$ to $m^2$ codes at a fixed embedding dimension, with closed-form correlations between any two
codes.

\section{Related work}
\label{sec:related}

\textbf{Interactive video world models.} World models predict the consequences of actions for
control~\citep{ha2018worldmodels,hafner2025dreamerv3}. Video world models learn this prediction
through autoregressive tokens~\citep{wu2024ivideogpt}, latent actions~\citep{bruce2024genie},
adapted video diffusion~\citep{huang2026vid2world}, and diffusion game
engines~\citep{alonso2024diamond,valevski2024gamengen,kanervisto2025wham}, and now generate
gameplay, explorable scenes, and navigation views under user
control~\citep{che2025gamegenx,li2025hunyuangamecraft,ye2025yan,mao2025yume,sun2025worldplay,
xiong2026actworld,bar2025navigationworldmodels}. Minecraft is a central testbed for
action-conditioned simulation~\citep{guo2025mineworld,yu2025gamefactory,zhang2025matrixgame,
he2025matrixgame2,wang2026matrixgame3} and for learned
agents~\citep{guss2019minerl,fan2022minedojo,baker2022vpt,lifshitz2023steve1}. Language models
write executable skills~\citep{liang2022codeaspolicies,wang2023voyager} and vision-language models
judge task success~\citep{du2023successdetectors}, and \modelname{} uses both to generate and
validate multiplayer interactions.

\textbf{Multiplayer world models.} Multiplayer models render every player's observation under
the combined actions of all players. Solaris, Gamma-World, and MIRA generate synchronized player
views in Minecraft and Rocket League~\citep{savva2026solaris,liu2026gammaworld,hu2026mira}, and
further models coordinate views through shared world states, registers, or
memories~\citep{wu2026multiworld,mo2026worldweaver,po2026multigen,zhao2026khora,cai2026mass}.
Related work learns multi-agent generation from single-view footage~\citep{hu2026metaworld},
routes peer-view evidence~\citep{xu2026consistworld}, composes joint actions for
cooperation~\citep{zhang2025combo}, and synthesizes surround views of a single
vehicle~\citep{wang2024drivewm,wen2024panacea,russell2025gaia2,cheng2026m4world}. ShareVerse and
Prisma-World widen coverage by stitching directional views or by cropping panoramic
recordings~\citep{zhu2026shareverse,sun2026prismaworld}. Their generated streams remain
perspective videos, and their multiplayer data come from live game
clients~\citep{savva2026solaris}. Like procedural generators that render programmable scenes
offline~\citep{greff2022kubric,deitke2022procthor}, \modelname{} renders recorded gameplay into
aligned POV and panoramic videos, and it predicts both.

\textbf{Panoramic video generation and world models.} Panoramic video records the full
surroundings in every frame, and recent world models adopt it as the world representation
itself. Panoramas serve as proxies for explorable 3D
worlds~\citep{hunyuanworld2025,yang2025matrix3d,su2026geniesim,wang2025evoworld,zhou2026moverse},
for long-horizon and single-image
exploration~\citep{liu2026omniroam,gui2025imageworld,lu2024genex}, as latent states for
perspective rendering~\citep{tan2026alayavista} and driving~\citep{li2026omninwm}, and for
geometric consistency and spatial
memory~\citep{jiang2026panoworld,li2026panoworldmemory,jia2026panoworld}. These models build on
360-degree video diffusion~\citep{wang2024360dvd,zhang2025panodit,xia2025panowan,ji2025campvg,
zhou2025holotime}, perspective-to-panorama
expansion~\citep{ma2024vidpanos,luo2025beyondframe,li2026cubecomposer}, and joint perspective and
panoramic modeling~\citep{tang2023mvdiffusion,zhang2024panfusion,tan2024imagine360,
fang2025viewpoint}. Perspective models instead recall regions from visual
history~\citep{xiao2025worldmem,huang2025memoryforcing} or simulate out-of-sight
dynamics~\citep{duan2026liveworld} for a single observer. \modelname{} extends panoramic
prediction to a shared world, where each panorama reflects changes caused by co-players.

\section{Panoramic multiplayer dataset}
\label{sec:data}

\begin{figure}[t]
\centering
\includegraphics[width=\linewidth]{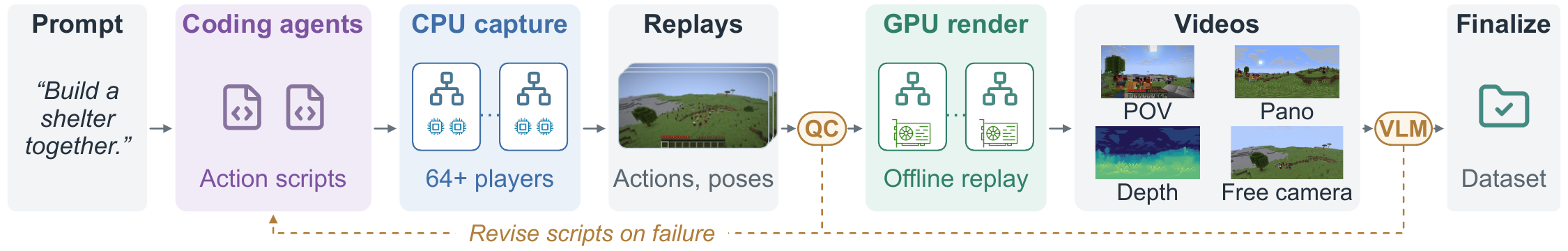}
\caption{\textbf{Distributed data generation.} Coding agents write action scripts, CPU capture
records sessions with more than 64 players without graphics clients, and GPU workers render the
replays offline. Behavioral QC and VLM review return failed scripts for revision.}
\label{fig:collection-overview}
\end{figure}

A panoramic multiplayer world model learns from both projections and the actions of every
player within one synchronized interaction. To this end, we construct a dataset that pairs each
player's POV video with a world-stabilized panoramic video, aligned actions and camera
annotations. Existing multiplayer data such as the Solaris dataset are recorded from live
Minecraft clients~\citep{savva2026solaris}, which limits each recording to the perspective view
displayed during play. Our pipeline instead records programmable interactions on CPUs and renders
the saved replays offline on GPUs, which yields POV and panoramic videos, relative depth and
custom camera trajectories from a single interaction without repeating the gameplay
(Figure~\ref{fig:collection-overview}).

\textbf{Scaling multiplayer data collection.} SolarisEngine pairs each Mineflayer controller
with a live graphics client~\citep{savva2026solaris}, and controller computation and rendering
therefore both grow with the player count and must keep pace with the running simulation.
\modelname{} removes rendering from gameplay. Its player runtime, built on the
open-source Azalea library, executes action scripts written in Rust and simulates interactions
among more than 64 simultaneous players while recording per-tick actions, positions, camera
orientations and player state. Without graphics clients, independent episodes run in parallel
across CPU hosts, and GPU workers render the saved replays on separate hosts, with views and
camera paths assigned to workers independently. Capture and rendering thus scale with separate
resources, and a large session is recorded once and rendered afterwards.

\textbf{Generating and validating interactions.} Interaction scenarios are specified as
natural-language descriptions of player behaviors and expected outcomes. Coding agents
compose reusable interaction primitives into executable scripts, which may fail to produce their
intended interaction, and each scenario is therefore validated on both behavior and appearance.
Behavioral quality control (QC) detects inactivity, ineffective movement and abrupt control
changes in the recorded actions and states before rendering, and a vision-language model (VLM)
then reviews sampled videos against the specified outcomes. Both checks return feedback
for the coding agents to revise the scripts.

\textbf{Observing a shared world beyond the local view.} Each player's rectilinear POV video is
paired with an equirectangular panorama rendered from the same camera origin. The
panorama stays oriented to the world as the player's head turns, which gives every frame a stable
directional reference and records changes that no local camera observes. All videos share the
server clock, and action annotations and camera poses bind each observation to
its player (Appendix~\ref{app:data-details}).

\section{Panoramic multiplayer world modeling}
\label{sec:method}

\begin{figure}[t]
\centering
\includegraphics[width=\linewidth]{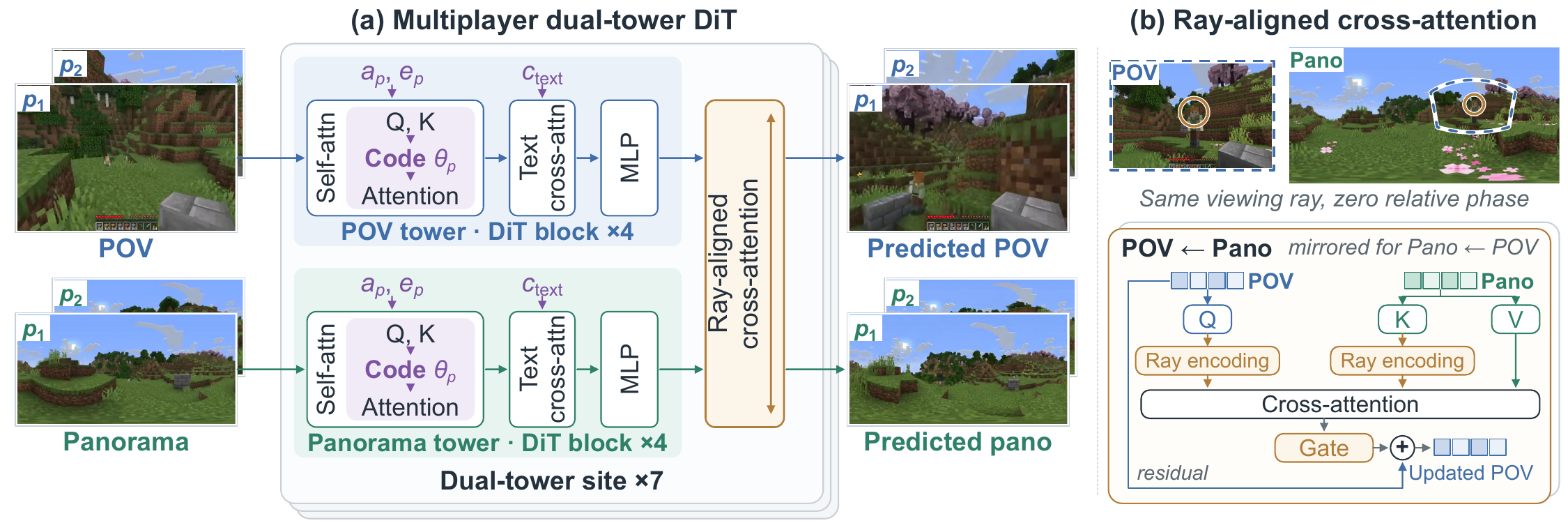}
\caption{\textbf{\modelname{} architecture.}
(a) POV and panorama towers share weights across players and are coupled by ray-aligned
cross-attention at seven sites. Actions $a_p$ and appearance attributes $e_p$ condition
self-attention, and the player code $\theta_p$ rotates Q and K. (b) Tokens that share a viewing
ray, shown by the POV field of view projected onto the panorama. Each direction rotates the
normalized Q and K by their viewing rays and adds its output through a zero-initialized gate.}
\label{fig:model-overview}
\end{figure}

The \modelname{} world model jointly predicts every player's POV and panoramic video under the
actions of all players (Figure~\ref{fig:model-overview}). We formulate the task
(Section~\ref{sec:task}) and describe the dual towers (Section~\ref{sec:two-tower}), the player
identity encoding (Section~\ref{sec:chirp}), the training objective (Section~\ref{sec:appearance})
and the causal few-step generator used for rollout (Section~\ref{sec:training}).

\subsection{Problem formulation}
\label{sec:task}

For $N$ players, the dataset provides synchronized POV and world-stabilized panoramic videos
with aligned actions. Let $p$ index players, $t$ observation time, $T$ the prediction horizon,
and $v\in\{\mathrm{pov},\mathrm{pano}\}$ the projection. Given the initial observations of all
players in both projections, \modelname{} models
\begin{equation}
p_\theta\!\left(\{X^v_{p,1:T}\}_{p,v}\,\middle|\,
\{X^v_{p,0}\}_{p,v},\{a_{p,0:T-1},e_p\}_{p=1}^N,c\right),
\label{eq:joint-task}
\end{equation}
where $X^v_{p,t}$ is an observation, $a_{p,t}$ contains the keyboard and camera controls of the
following transition, and $e_p$ is a fixed appearance attribute of the player's character. The
shared condition $c$ contains a scene description and the initial camera orientations, from
which the angular controls determine every later orientation. Future positions, visibility and
images are never provided (Appendix~\ref{app:conditioning}).

\subsection{Dual towers with ray-aligned cross-attention}
\label{sec:two-tower}

\textbf{Dual-tower backbone.} POV and panoramic frames describe the same interaction through
different projections. \modelname{} builds on the DiT backbone of
Cosmos-Predict2.5~\citep{nvidia2025cosmospredict25} and uses two towers of identical
architecture, one predicting the POV videos and the other the panoramic videos, following a
similar design of perspective and panoramic diffusion
models~\citep{zhang2024panfusion,tan2024imagine360}. Each tower shares its weights across
players. A tower encodes the initial frames into latent context, combines it with noisy future
latents, and processes the result with DiT blocks of self-attention, text cross-attention and an
MLP. The actions $a_{p,t}$ of each player
condition self-attention in both towers, together with the appearance attribute $e_p$ that
Section~\ref{sec:chirp} motivates. The panorama tower further accounts for the spherical geometry
of the equirectangular grid. A token in latent row $i$ of $H$ rows and column $k$ receives rotary
position coordinates
\begin{equation}
\big(y_i,\,k\big),\qquad
y_i=\frac{H-1}{2}\big(1-\sin\vartheta_i\big),\qquad
\vartheta_i=\pi\Big(\frac12-\frac{i}{H-1}\Big),
\label{eq:pano-rope}
\end{equation}
where $\vartheta_i$ is the latitude of row $i$. The vertical coordinate thus follows the
equal-area sine of latitude and compresses the rows near the poles, which the equirectangular
projection oversamples, a design adopted from PanoWorld~\citep{jiang2026panoworld}. Writing
$\mathbf h^v_l$ for the tokens of tower $v$ after block $\mathcal B^v_l$ and $\mathcal X_l$ for a
bidirectional ray-aligned cross-attention module, the two towers run in parallel as
\begin{equation}
\mathbf h^v_l=\mathcal B^v_l\big(\mathbf h^v_{l-1};\,s,\{a_{p,t},e_p\},c\big),\qquad
\big(\mathbf h^{\mathrm{pov}}_l,\mathbf h^{\mathrm{pano}}_l\big)\leftarrow
\mathcal X_l\big(\mathbf h^{\mathrm{pov}}_l,\mathbf h^{\mathrm{pano}}_l\big)
\ \ \text{for}\ l\in\{4,8,\dots,L\},
\label{eq:dual-tower}
\end{equation}
where $s$ is the noise level, $L$ the number of blocks, and each such $l$ marks a
cross-attention site. In each $\mathcal X_l$, the POV and
panoramic tokens of one player at one latent time attend to each other, each located by its
world-frame viewing ray as detailed below. We initialize the shared backbone weights of the POV tower from Gamma-World~\citep{liu2026gammaworld} and those of the panorama
tower from PanoWorld~\citep{jiang2026panoworld}, and zero-initialize the cross-attention gates and the
weights of all new input features.

\textbf{World-frame viewing rays.} \modelname{} locates every token by the world-frame
direction it observes. A POV token at image coordinates $(u,v)$ receives the unit ray
$\mathbf d^{\mathrm{pov}}_{p,t}=O_{p,t}\operatorname{normalize}(K^{-1}[u,v,1]^{\mathsf T})$, where
$K$ holds the POV intrinsics and $O_{p,t}$ the camera orientation given by the yaw and pitch of
player $p$ at time $t$. A panorama token at longitude $\varphi$ and latitude $\vartheta$ receives
$\mathbf d^{\mathrm{pano}}=(\cos\vartheta\sin\varphi,\sin\vartheta,\cos\vartheta\cos\varphi)$
in the world frame (Appendix~\ref{app:geometry}). Following the use of camera geometry in
multi-view attention~\citep{li2025prope,wu2026rayrope,xiang2026viewrope}, each ray is mapped to
rotary phases $\Phi(\mathbf d)=(\omega_j d_x,\omega_j d_y,\omega_j d_z)_{j=1}^{J}$ with $J$
frequencies per Cartesian component, and $\mathsf R(\boldsymbol\phi)$ denotes the block-diagonal
rotation by these phases.

\textbf{Ray-aligned cross-attention.} The module $\mathcal X_l$ connects both views of one player at one
latent time, for which the two camera centers coincide. For a POV query $\mathbf q$ and a panorama key
$\mathbf k$, rotating each by its own ray and applying the relative-phase property of rotary
encoding~\citep{su2021roformer} gives
\begin{equation}
\big(\mathsf R(\Phi(\mathbf d^{\mathrm{pov}}))\mathbf q\big)^{\mathsf T}
\mathsf R(\Phi(\mathbf d^{\mathrm{pano}}))\mathbf k
=\mathbf q^{\mathsf T}
\mathsf R\!\left(\Phi(\mathbf d^{\mathrm{pano}})-\Phi(\mathbf d^{\mathrm{pov}})\right)
\mathbf k.
\label{eq:ray-relative}
\end{equation}
Tokens that observe the same direction thus have zero relative phase regardless of their image
coordinates and of the head orientation, and this phase correspondence holds by construction at
every frame. Writing $\mathrm{Attn}_{\mathrm{ray}}$ for attention with Q and K normalized per head and
rotated by their rays, the POV features $\mathbf h^{\mathrm{pov}}$ read the panoramic features
$\mathbf h^{\mathrm{pano}}$ through
\begin{equation}
\mathbf h^{\mathrm{pov}}\leftarrow\mathbf h^{\mathrm{pov}}
+\mathbf g\odot W_o\,\mathrm{Attn}_{\mathrm{ray}}\big(W_q\mathbf h^{\mathrm{pov}},
W_k\mathbf h^{\mathrm{pano}},W_v\mathbf h^{\mathrm{pano}}\big),
\label{eq:exchange}
\end{equation}
and the reverse direction mirrors this update with its own projections and gate. The channel
gate $\mathbf g$ starts from zero, and temporal and cross-player attention remain within each
tower.

\subsection{Player identity encoding}
\label{sec:chirp}

\textbf{Rotary player codes.} A model that shares parameters across players, as set models do
for inputs of variable size~\citep{zaheer2017deepsets,lee2018settransformer}, must associate each
observation stream with its own actions for any population. Following
Gamma-World~\citep{liu2026gammaworld}, \modelname{} reserves $m$ rotary phase planes in
self-attention for player identity and rotates the queries and keys of player $p$
by a code $\boldsymbol\theta_p\in[0,2\pi)^m$. In the attention between players $p$ and $p'$,
this band contributes
\begin{equation}
\mathrm{Re}\sum_{r=0}^{m-1}q_r\,\bar k_r\,
e^{\mathrm i(\theta_{p,r}-\theta_{p',r})},
\label{eq:player-phase}
\end{equation}
where $q_r$ and $k_r$ are the complex feature pairs of plane $r$. The relative code
$\boldsymbol\theta_p-\boldsymbol\theta_{p'}$ therefore enters the attention of every token
pair in each head. Gamma-World places the codes at the vertices of a regular simplex,
which admits at most $m+1$ equidistant codes in $m$ phase planes.

\textbf{Quadratic-phase codes.} We draw instead on quadratic-phase codes from mutually unbiased
bases~\citep{wootters1989optimal,klappenecker2003mub}. For an odd prime $m$,
\begin{equation}
\theta^{(a,b)}_r=\frac{2\pi}{m}(ar^2+br),\qquad
z^{(a,b)}_r=\frac{1}{\sqrt m}e^{\mathrm i\theta^{(a,b)}_r},\qquad
a,b,r\in\mathbb F_m,
\label{eq:chirp}
\end{equation}
where $\mathbb F_m$ is the field of integers modulo $m$, $a$ indexes a code family and $b$ a code
within that family. The normalized inner products of the $m^2$ codes have magnitudes
\begin{equation}
\left|\left\langle z^{(a,b)},z^{(a',b')}\right\rangle\right|
=\begin{cases}
1,&(a,b)=(a',b'),\\
0,&a=a',\ b\ne b',\\
m^{-1/2},&a\ne a'.
\end{cases}
\label{eq:chirp-correlation}
\end{equation}
Codes within a family are exactly orthogonal by finite geometric sums, and codes across families
have the uniform correlation $m^{-1/2}$ given by quadratic Gauss sums
(Appendix~\ref{app:chirp-proof}). After reserving the all-zero code, \modelname{} provides
$m^2-1$ codes, compared with $m+1$ simplex codes, and the identity pool thus grows
quadratically rather than linearly with the embedding dimension.

\textbf{Randomized code assignment.} Each player keeps one code across time and projections.
Codes are resampled per training example, and observations, actions, attributes and codes are
permuted together. The model thus distinguishes players by relative codes rather than indices,
and the codebook serves populations up to its size without per-player parameters
(Appendix~\ref{app:permutation}).

\textbf{Appearance attributes.} Because code assignments change across examples, a code
identifies a player stream but not a character. The appearance attribute $e_p$ specifies the
character appearance and follows its player under permutation. It is concatenated with the camera
controls $a^{\mathrm{cam}}_{p,t}$, which contain the yaw and pitch increments and the absolute
orientation, to form the camera condition $[a^{\mathrm{cam}}_{p,t},e_p]$ in both towers.

\subsection{Training objective}
\label{sec:appearance}

\textbf{Joint flow matching.} Both towers are trained with flow
matching~\citep{lipman2023flowmatching}. For clean latents $Z^v$ of projection $v$, independent
Gaussian noise $\epsilon^v$ and a noise level $s$ shared by the two projections, the model
$F_\theta$ predicts the velocity $U^v=\epsilon^v-Z^v$ at $Z^v_s=(1-s)Z^v+s\epsilon^v$ given the
conditions $\mathcal C$ of Equation~\ref{eq:joint-task}, and the joint objective sums the losses of the two
projections,
\begin{equation}
\mathcal L=\mathcal L_{\mathrm{pov}}+\mathcal L_{\mathrm{pano}},\qquad
\mathcal L_v=\mathbb E\Big[\frac{1}{|\Omega_v|}\sum_{i\in\Omega_v}\bar w_i^v
\big\|F^v_{\theta,i}(Z_s,s,\mathcal C)-U^v_i\big\|_2^2\Big],
\label{eq:joint-loss}
\end{equation}
where $\Omega_v$ indexes the latent positions of all players and forecast frames in a sample.

\textbf{Co-player weighting.} Co-players often occupy a small part of the frame, and an
image-wide loss therefore underweights their appearance and motion. Projecting the recorded body
boxes of the co-players into each view gives a support mask $M_i\in\{0,1\}$, and the position
weights
\begin{equation}
\bar w_i=\frac{1+\lambda M_i}{1+\lambda\bar M},\qquad
\bar M=\frac{1}{|\Omega_v|}\sum_{i\in\Omega_v}M_i,
\label{eq:coplayer-weight}
\end{equation}
raise the loss on co-player regions while keeping unit mean per sample. Half of the training
windows further favor co-player occupancy to raise exposure to interactions
(Appendix~\ref{app:foreground}).

\subsection{Causal generation and distillation}
\label{sec:training}

\textbf{Block-causal generation.} The two pretrained towers are first adapted to their own
projections, after which joint training with the cross-attention modules minimizes
Equation~\ref{eq:joint-loss} and yields a bidirectional teacher. For rollout, a causal generator
uses diffusion
forcing~\citep{chen2024diffusionforcing} with the block-causal sparse hub attention of
Gamma-World~\citep{liu2026gammaworld}. Within each tower, players attend to their own histories
and to shared hub tokens, which mediate cross-player interaction. Noise levels vary across temporal
blocks and are shared by all players and both projections within a block. A query in block $b$
accesses only blocks $b'\leq b$, which the same-time cross-attention preserves.

\textbf{Few-step distillation.} We distill the causal generator $G_\phi$ on its own rollouts with
Self Forcing and distribution matching distillation (DMD)~\citep{huang2025selfforcing,yin2024dmd}.
A frozen bidirectional teacher provides the real score $s_{\mathrm{real}}$, and a learned
bidirectional network tracks the score $s_{\mathrm{fake}}$ of the generated distribution. Both
start from the joint teacher and receive identical conditions and player codes. For a generated
latent $\hat Z=G_\phi(\cdot)$ re-noised to $\hat Z_s$, the generator follows the co-player
weighted gradient
\begin{equation}
\nabla_\phi\mathcal L_{\mathrm{DMD}}
=\mathbb E\Big[\sum_i\bar w_i\big(s_{\mathrm{fake}}(\hat Z_s,s)-s_{\mathrm{real}}(\hat Z_s,s)\big)_i
\frac{\partial\hat Z_i}{\partial\phi}\Big].
\label{eq:dmd}
\end{equation}
Causal training and distillation sample $N\in\{2,4,8,16\}$ players, with shorter clips at larger
populations. At inference, the distilled model generates the two streams jointly block by block
with four denoising steps per block. Appendix~\ref{app:implementation} reports the training
lengths and sampling settings.

\section{Experiments}
\label{sec:exp}

\subsection{Experimental setup}

\textbf{Models and baselines.} We evaluate \modelname{} (Causal), sampled with 35 steps, and
\modelname{} (Distilled), sampled with four steps per causal block. The baselines are the
bidirectional, causal and distilled variants of Gamma-World~\citep{liu2026gammaworld},
Solaris~\citep{savva2026solaris}, and MultiWorld~\citep{wu2026multiworld}, which runs its
official 50-step autoregressive sampler with Minecraft actions mapped to its action space. No
prior multiplayer world model accepts panoramas, and the baselines thus receive only the initial
POV observations. \modelname{} also receives the initial panoramas, and it outperforms every
baseline even when its cross-attention is disabled at inference (Table~\ref{tab:inference-sites}, Appendix~\ref{app:interventions}).

\textbf{Protocol and metrics.} The main comparison uses 30 two-player interaction episodes of 12 seconds at
20\,Hz, and separate sets of 12 episodes cover four, eight and sixteen players. All evaluation
episodes are absent from the final training split (Appendix~\ref{app:evaluation}). We report PSNR, SSIM and LPIPS~\citep{zhang2018lpips} over each player's
239 forecast frames, together with
FID~\citep{heusel2017fid} over all forecast frames and FVD~\citep{unterthiner2018fvd} over
16-frame clips.

\subsection{Comparison with multiplayer world models}

\begin{table}[t]
\caption{Two-player interaction tier with 30 episodes and 239 forecast frames per player. Baselines support only POV input, and PAWA additionally receives the initial panoramas. \textbf{Bold} and \underline{underline} mark the best and second best within each stream and evaluation group.}
\label{tab:comparison}
\centering\footnotesize\setlength{\tabcolsep}{3.5pt}
\begin{tabular}{@{}llS[table-format=3.1]S[table-format=3.2]S[table-format=1.3]S[table-format=2.2]S[table-format=1.3]@{}}
\toprule
Model & Stream & {FVD $\downarrow$} & {FID $\downarrow$} & {LPIPS $\downarrow$} & {PSNR $\uparrow$} & {SSIM $\uparrow$} \\
\midrule
Gamma-World (Bidirectional) & POV & 267.1 & 56.63 & 0.567 & 13.08 & 0.478 \\
Gamma-World (Causal) & POV & 306.7 & 52.00 & 0.606 & 12.46 & 0.451 \\
Gamma-World (Distilled) & POV & 279.7 & 58.36 & 0.584 & 11.83 & 0.431 \\
Solaris & POV & 614.9 & 77.31 & 0.634 & 10.42 & 0.447 \\
MultiWorld & POV & 771.2 & 186.06 & 0.667 & 11.76 & 0.395 \\
PAWA (Causal) & POV & \bfseries 120.1 & \snd{000.00}{30.80} & \bfseries 0.421 & \bfseries 16.86 & \bfseries 0.562 \\
PAWA (Causal) & Panorama & \bfseries 88.0 & \bfseries 26.53 & \bfseries 0.223 & \bfseries 20.41 & \bfseries 0.643 \\
PAWA (Distilled) & POV & \snd{000.0}{138.1} & \bfseries 28.40 & \snd{0.000}{0.465} & \snd{00.00}{15.38} & \snd{0.000}{0.516} \\
PAWA (Distilled) & Panorama & \snd{000.0}{98.1} & \snd{000.00}{28.26} & \snd{0.000}{0.243} & \snd{00.00}{19.77} & \snd{0.000}{0.617} \\
\bottomrule\end{tabular}
\end{table}

\textbf{Prediction quality.} Both \modelname{} variants surpass every baseline on every POV
metric of Table~\ref{tab:comparison}. \modelname{} (Causal) more than halves the POV FVD of the
strongest baseline, Gamma-World (Bidirectional), and \modelname{} (Distilled) does so against
Gamma-World (Distilled) under the same four-step budget. The panoramic predictions, which no prior
multiplayer model provides, reach high fidelity as well. Co-players entering a view after at least 1\,s
outside it reach LPIPS 0.58 against 0.65 for the best baseline, and the panorama renders
co-players outside the POV as well as inside it
(Appendix~\ref{app:out-of-view}).

\begin{figure}[t]\centering
\includegraphics[width=\linewidth,trim=28 0 0 0,clip]{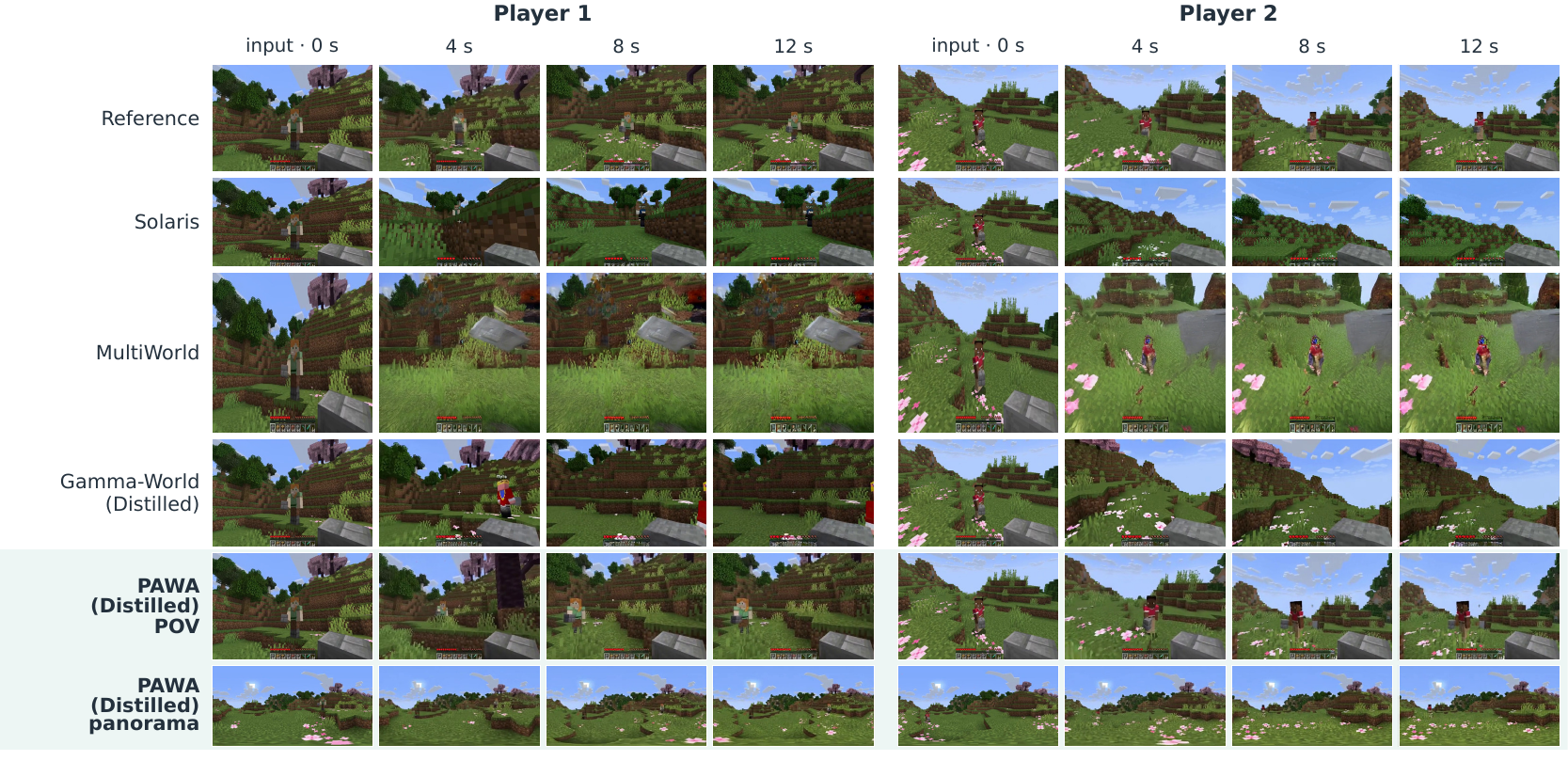}
\caption{\textbf{Synchronized two-player forecasts.} Reference denotes the recorded
continuation, and POV and panorama rows share time columns. The co-player in each POV can be compared with
the one in the Reference row. MultiWorld uses 50 denoising steps per block and the other models use
four.}
\label{fig:rollout}\end{figure}

\begin{figure}[t]
\centering
\includegraphics[width=\linewidth]{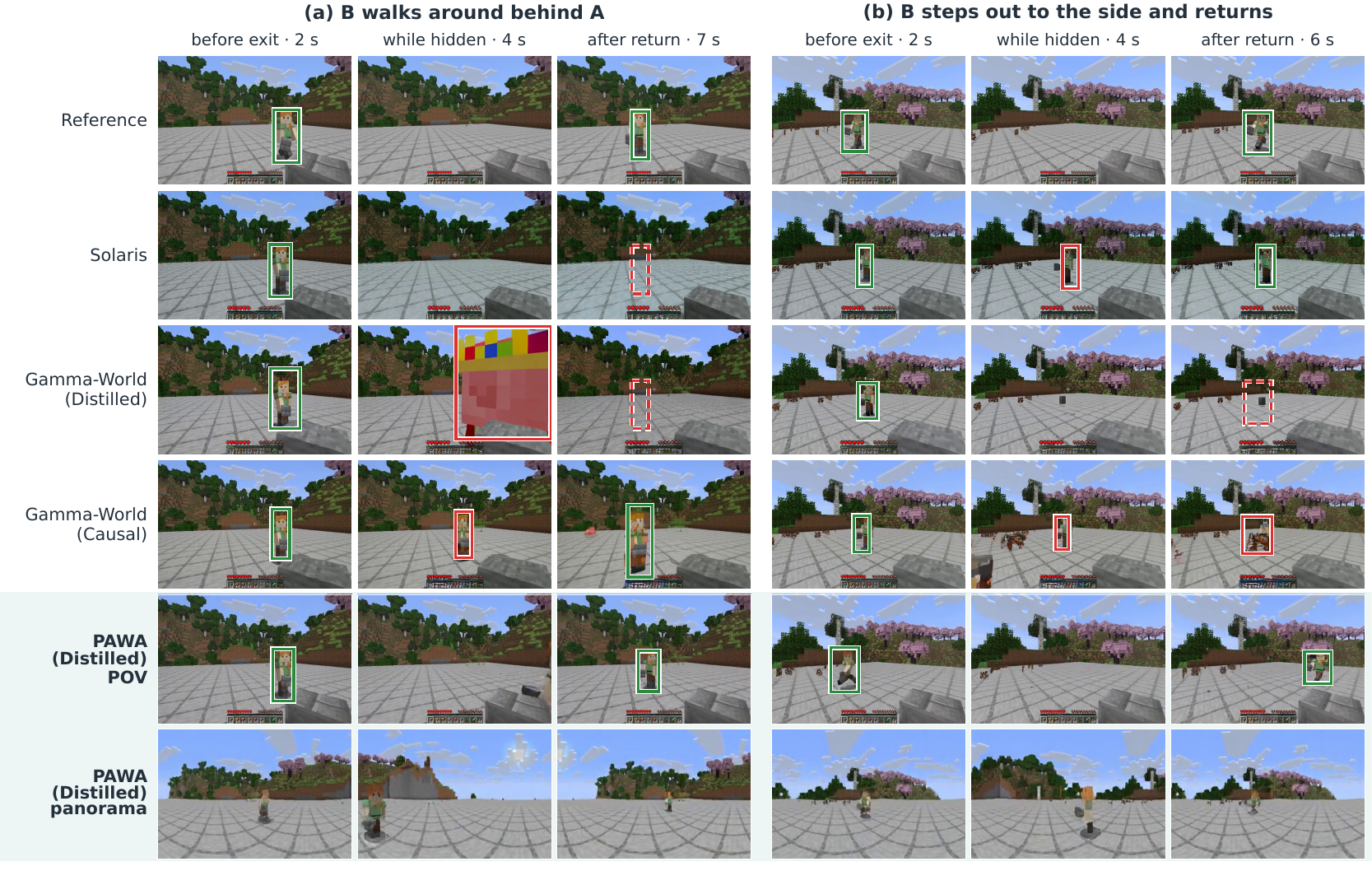}
\caption{\textbf{A co-player leaving and re-entering the view.} In (a) the co-player walks
around behind the player, and in (b) it steps out to the side and returns. Green boxes mark a correct
co-player, red boxes a co-player that should be hidden or a wrong character, and dashed red boxes
a co-player the model has lost. The baselines keep the co-player in view, replace it or lose
it, whereas the panorama of \modelname{} (Distilled) tracks the hidden co-player and its POV
shows it again after the return.}
\label{fig:reentry}
\end{figure}

\textbf{Qualitative comparison.} In Figure~\ref{fig:rollout}, Gamma-World (Distilled) replaces
the co-player of Player~1 with a differently dressed character and loses the co-player of
Player~2 after 4\,s, Solaris drops the co-player of Player~2, and MultiWorld stalls after 4\,s,
whereas \modelname{} keeps both co-players in view with a consistent appearance throughout the
12-second rollout. In
Figure~\ref{fig:reentry}, where a co-player leaves and re-enters the view, \modelname{} renders its
return while the baselines lose it or wrongly keep it in view.

\subsection{Scaling the player population}

\begin{table}[t]
\caption{POV prediction with larger populations. \textbf{Bold} and \underline{underline} mark
the best and second best.}
\label{tab:population-pov}
\centering\footnotesize\setlength{\tabcolsep}{4pt}
\begin{tabular}{@{}lS[table-format=3.1]S[table-format=1.3]S[table-format=4.1]S[table-format=1.3]S[table-format=3.1]S[table-format=1.3]@{}}
\toprule
& \multicolumn{2}{c}{4 players} & \multicolumn{2}{c}{8 players} & \multicolumn{2}{c}{16 players} \\
\cmidrule(lr){2-3}\cmidrule(lr){4-5}\cmidrule(l){6-7}
Model & {FVD $\downarrow$} & {LPIPS $\downarrow$} & {FVD $\downarrow$} & {LPIPS $\downarrow$} & {FVD $\downarrow$} & {LPIPS $\downarrow$} \\
\midrule
Gamma-World (Bidirectional) & 752.3 & 0.633 & 1707.7 & 0.681 & {--} & {--} \\
Gamma-World (Causal) & 373.0 & 0.587 & 297.3 & 0.606 & {--} & {--} \\
Gamma-World (Distilled) & 323.6 & 0.577 & 267.1 & 0.586 & {--} & {--} \\
PAWA (Causal) & \bfseries 138.0 & \bfseries 0.427 & \bfseries 111.5 & \bfseries 0.458 & \bfseries 92.2 & \bfseries 0.467 \\
PAWA (Distilled) & \snd{000.0}{167.3} & \snd{0.000}{0.481} & \snd{0000.0}{141.3} & \snd{0.000}{0.517} & \snd{000.0}{116.7} & \snd{0.000}{0.520} \\
\bottomrule
\end{tabular}
\end{table}

In Table~\ref{tab:population-pov}, the released Gamma-World configuration provides four simplex
player codes, and an eight-code extension enables it to run with eight players, yet its bidirectional
variant collapses and its best variant has more than twice the POV FVD of \modelname{} (Causal). With sixteen players, which
only \modelname{} supports, every player receives a distinct quadratic-phase code, and
for 73 to 83\% of the players whose camera turns, their own actions best explain the rotation of their stream, against 6\%
by chance (Appendix~\ref{app:out-of-view}).

\subsection{Ablations}

\textbf{Ray-aligned cross-attention.} Disabling all seven sites of \modelname{}
(Distilled) at inference raises POV FVD
from 138.1 to 179.0 and LPIPS from 0.465 to 0.510 (Table~\ref{tab:inference-sites}). The
pano$\to$POV direction alone accounts for the entire POV degradation,
and early and late sites contribute comparably. POV
prediction thus draws on the panorama.
Continued for 3000 updates from the same checkpoint, the causal model without cross-attention
sites reaches POV LPIPS 0.511 and FVD 191.7 against 0.455 and 138.4 with them, and the LPIPS gap grows from 0.012 at the start
(Appendix~\ref{app:train-ablation}).

\textbf{Learned ray alignment.} At the late sites, POV queries place up to 57\% of their
attention on the panorama inside their own frustum, against 10\% for uniform attention and for
re-initialized sites (Table~\ref{tab:ray-alignment}). The cross-attention thus learns to connect
matching directions.

\FloatBarrier
\section{Conclusion}
\label{sec:conclusion}

We present \modelname{}, a dataset and framework for action-conditioned panoramic multiplayer
world modeling. Its pipeline turns sessions of more than 64 players into
synchronized POV and world-stabilized panoramic videos with aligned actions, and its model
jointly predicts both videos for every player through ray-aligned cross-attention and
quadratic-phase player codes. It outperforms state-of-the-art multiplayer world models in POV
prediction and produces high-fidelity panoramas.

\textbf{Limitations and future work.} Like all evaluated multiplayer world models, \modelname{}
loses co-player fidelity over long rollouts, and its POV and panoramic forecasts diverge gradually
since the cross-attention aligns directions without enforcing pixel agreement. Although the
pipeline records more than 64 players and the codebook provides 120 identities, the model does not
decouple the players, and its computational cost increases rapidly with the number of players,
which limits training and evaluation to sixteen players. Scaling to larger populations requires a
redesign in which each player runs on its own GPU and exchanges little information with the others.

\clearpage

\bibliographystyle{plainnat}
\bibliography{refs}

\clearpage
\setcounter{topnumber}{3}
\setcounter{totalnumber}{4}
\linespread{1.05}\selectfont
\beginappendix
\section{Dataset specification}
\label{app:data-details}

\begin{table}[!ht]
\caption{Collection snapshot. Player views count complete per-player renders, each carrying a POV
stream, a panorama stream and a per-frame action record.}
\label{tab:d1}
\begin{center}
\begin{tabular}{lS[table-format=4]S[table-format=5]S[table-format=3]S[table-format=4]S[table-format=4]S[table-format=4]S[table-format=4]}
\toprule
 & {Episodes} & {Player views} & \multicolumn{5}{c}{Episodes by player count} \\
\cmidrule(l){4-8}
 & & & {$N{=}1$} & {$N{=}2$} & {$N{=}4$} & {$N{=}8$} & {$N{=}16$} \\
\midrule
Collection snapshot & 8162 & 45716 & 350 & 2555 & 2770 & 1327 & 1160 \\
\bottomrule
\end{tabular}
\end{center}
\end{table}

Table~\ref{tab:d1} summarizes the collection. A player view contains both projections and the
per-frame action record of one player and is counted once. The source videos have POV resolution
$640\times360$ and panorama resolution $1024\times512$ at 20\,Hz. Every stream is mapped to the
world-age ticks of the server and cropped to the common interval covered by the actions of all
retained players, and a three-tick display-phase correction aligns actions with the rendered
frames. Each replay also exports relative depth in both projections and can be rendered again
along custom camera trajectories within the recorded scene. The collector has recorded sessions
with 64 and 128 simultaneous players, from which participant and observer views are rendered on
demand.

\section{Model and training settings}
\label{app:implementation}

\subsection{Conditioning and architecture}
\label{app:conditioning}
The camera condition contains the yaw and pitch increments and scaled sine and cosine pairs of the
absolute yaw and pitch, followed by a 16-dimensional appearance attribute, 22 dimensions in total.
Keyboard actions form a separate input. Absolute orientations follow from the initial orientation
and the angular controls, and a recorded angular increment is reindexed as the outgoing
transition of Equation~\ref{eq:joint-task}. Both projections share the world clock, the player
ordering and the player codes.

Each tower has 28 blocks of width 2048 with 16 heads, and the cross-attention sites follow blocks
$3,7,\ldots,27$ (zero-based). The panorama tower uses the equirectangular rotary position encoding of Equation~\ref{eq:pano-rope}. Each causal tower has eight hub tokens per latent frame.

The player-code band has $m=11$ phase planes, which provide 120 assignable codes after reserving
the all-zero code. For $2\leq N\leq11$ players, the codes come from a single family with probability
0.75 and from several families otherwise, and larger populations always span several families.
The ray encoding allocates 16 phase planes to each Cartesian component with frequencies
$\omega_j=32^{j/15}$, $j=0,\ldots,15$, and uses the POV field of view of
$102.45^\circ\times70^\circ$.

\subsection{Objectives and window sampling}
\label{app:foreground}
The flow-matching loss of Equation~\ref{eq:joint-loss} additionally applies the base noise
weighting of the backbone and excludes the supplied context frames. Joint bidirectional training
shares one noise level per sample, and causal training shares one level across players and
projections within each temporal block. Appearance adaptation and joint training use the
projected co-player support with $\lambda=4$. The causal and distilled models expand the body
boxes by $1.4\times1.2$, take their union over lags of 0 to 6 ticks to absorb small alignment
offsets, and use
\begin{equation}
\lambda=\min\!\left(2,\frac{0.25}{\bar M+10^{-8}}\right),
\label{eq:weight-budget}
\end{equation}
which keeps the background weight at least 0.8. Distillation applies these weights to both the
generator and the learned score. Half of the training windows are sampled uniformly, and the
other half are selected from four candidates with probability proportional to $0.1+f_j$, where $f_j$
is the co-player frustum occupancy. Joint training samples $N\in\{1,2,4,8\}$ players, and causal
training and distillation sample $N\in\{2,4,8,16\}$, weighted toward smaller populations.

\subsection{Optimization}
\label{app:transitions}
For the bidirectional teacher, the POV and panorama towers each receive 10000 updates on their
own projection, followed by appearance adaptation and 10000 joint updates on 93-frame clips at
POV resolution $480\times320$ and panorama resolution $512\times256$, with AdamW at learning rate
$3\times10^{-5}$, betas $(0.9,0.99)$ and weight decay 0.001. No classifier-free guidance is used.
The causal generator starts from block-causal models trained for 15000 updates and is trained
for a further 9000 updates with quadratic-phase codes at learning rate $10^{-5}$, on clips of
93 frames for up to eight players and 45 frames for sixteen. Distillation initializes the
generator from the causal model and both scores from the bidirectional teacher, and alternates
one generator update with four learned-score updates for 2000 generator updates, with AdamW
learning rates $2\times10^{-6}$ and $4\times10^{-7}$, betas $(0,0.999)$ and weight decay 0.01,
on clips of 45, 45, 33 and 21 frames for 2, 4, 8 and 16 players. Three latent frames form a
causal block, and the four denoising steps use timesteps $(1000,937,833,625)$. Training uses
BF16 computation with FP32 parameters and optimizer states.

\section{Geometry and player-code analysis}
\label{app:theory}

\subsection{Viewing rays}
\label{app:geometry}
For normalized image coordinates $(u,v)$, camera-to-world orientation $O_{p,t}$, and horizontal
and vertical POV fields of view $\alpha_x,\alpha_y$, the paired world-frame rays are
\begin{equation}
\begin{split}
\mathbf d^{\mathrm{pov}}_{p,t}(u,v)
 &=O_{p,t}\operatorname{normalize}\!\begin{pmatrix}
 (2u-1)\tan(\alpha_x/2)\\ (1-2v)\tan(\alpha_y/2)\\1
 \end{pmatrix},\\
\mathbf d^{\mathrm{pano}}(u,v)
 &=\begin{pmatrix}\cos\vartheta\sin\varphi\\\sin\vartheta\\
 \cos\vartheta\cos\varphi\end{pmatrix},\quad
\varphi=2\pi(u-\tfrac12),\quad\vartheta=\pi(\tfrac12-v).
\end{split}
\label{eq:world-rays}
\end{equation}
The camera axes are right, up and forward, and the panorama axes are fixed to the world. The
spherical map is continuous at the longitude seam. For block-diagonal rotary maps,
$\mathsf R(\phi)^{\mathsf T}\mathsf R(\psi)=\mathsf R(\psi-\phi)$, which gives
Equation~\ref{eq:ray-relative}. Matching rays therefore have zero relative phase for the
co-located POV and panoramic cameras of the same player.

\subsection{Simplex capacity and quadratic-phase correlations}
\label{app:chirp-proof}
For $V$ equidistant points in $\mathbb R^m$ with squared separation $\delta^2>0$, the centered
Gram matrix is $(\delta^2/2)(I_V-\mathbf1\mathbf1^{\mathsf T}/V)$, of rank $V-1$, and thus
$V\leq m+1$.

Let $\omega=e^{2\pi\mathrm i/m}$, $A=a'-a$, and $B=b'-b$ for the codes in
Equation~\ref{eq:chirp}. Their Hermitian inner product is
$m^{-1}\sum_r\omega^{Ar^2+Br}$. It equals one for $A=B=0$ and vanishes for $A=0,B\ne0$
by the finite geometric sum. For $A\ne0$, write $G(A,B)=\sum_r\omega^{Ar^2+Br}$ and set
$r=s+h$ to obtain
\begin{equation}
|G(A,B)|^2=\sum_h\omega^{Ah^2+Bh}\sum_s\omega^{2Ahs}=m.
\label{eq:gauss-proof}
\end{equation}
Because $m$ is an odd prime, $2A$ is invertible and only $h=0$ contributes. Division by $m$
gives the cross-family magnitude $m^{-1/2}$ in Equation~\ref{eq:chirp-correlation}, and the $m$
families of $m$ orthonormal codes together give $m^2$ codes in $m$ phase planes, against the
$m+1$ vertices of a regular simplex.

\paragraph{Player relabeling.}
\label{app:permutation}
Shared player-wise maps and same-player cross-attention commute with a joint permutation of
observations, actions, appearance attributes, codes, noise and masks. Random code assignment
preserves this property in distribution when its law is independent of player order, and the
evaluated models share block noise levels across players. The distribution of generated videos is
therefore equivariant to any reordering of the players, and no player index carries meaning.

\section{Evaluation protocol and supplementary results}
\label{app:evaluation}

\subsection{Protocol}

\textbf{Episodes.} The interaction tier contains 30 two-player episodes, one window per scenario
family, selected from seeded candidates by the geometric co-visibility of the players in the
recorded telemetry rather than by generated quality. The low-motion tier selects 12 further
two-player windows by camera motion, and the four-, eight- and sixteen-player tiers contain 12
episodes each. Six of the interaction episodes also served for checkpoint selection. On the
other 24 episodes, POV LPIPS is 0.418 for \modelname{} (Causal) and 0.459 for \modelname{}
(Distilled) against 0.558 for Gamma-World (Bidirectional), and the LPIPS ranking of
Table~\ref{tab:comparison} is unchanged.

\textbf{Metrics.} Scoring excludes the conditioning frame and covers 239 forecast frames per
player, ending at 11.95\,s. Solaris and MultiWorld generate 241 frames and the other models 249,
and the first 240 are decoded. PSNR uses the pixel MSE of the whole window, SSIM and AlexNet LPIPS
are averaged over the forecast frames, and MAE uses the 0 to 255 scale. Per-player scores are averaged across players and then episodes. Pixel metrics use the natively
preprocessed ground truth of each model. FID pools 2048-dimensional InceptionV3 features of all
forecast frames, and FVD pools I3D features of 14 disjoint 16-frame clips per player at
$224\times224$.

\textbf{Baselines.} The bidirectional and causal Gamma-World variants use 35 steps and the
distilled one uses four. Solaris uses its official
distilled four-step sampler with the aligned initial observation. MultiWorld uses its public
checkpoint and official autoregressive sampler with 50 steps and three 81-frame blocks, each
conditioned on the tail of the previous block, and its action space receives the movement, jump
and camera controls. The released Gamma-World configuration provides four player codes, and the
eight-player comparison extends its simplex player encoding to eight codes, whereas Solaris and
MultiWorld support two players.

\subsection{Additional comparisons}
\label{app:distillation}

Tables~\ref{tab:easy-full}, \ref{tab:four-player} and~\ref{tab:population} extend the comparison
to the low-motion, four-player, eight-player and sixteen-player tiers, where \modelname{} again
obtains the best POV scores. Distillation substantially improves the four-step FVD and LPIPS. Sampling the causal
model directly with the same four steps yields POV FVD 438.5 and LPIPS 0.497 on the interaction
tier, which distillation lowers to 138.1 and 0.465. Varying only the inference seed of
\modelname{} (Distilled) on an 18-episode subset changes POV FVD within 157.4 to 159.6 and POV
LPIPS within 0.468 to 0.475, far less than the gaps to the baselines.

\begin{table}[t]
\caption{Two-player low-motion tier with 12 episodes and 239 forecast frames per player. \textbf{Bold} and \underline{underline} mark the best and second best within each stream and evaluation group.}
\label{tab:easy-full}
\centering\footnotesize\setlength{\tabcolsep}{3.5pt}
\begin{tabular}{@{}llS[table-format=3.1]S[table-format=3.2]S[table-format=1.3]S[table-format=2.2]S[table-format=1.3]S[table-format=2.2]@{}}
\toprule
Model & Stream & {FVD $\downarrow$} & {FID $\downarrow$} & {LPIPS $\downarrow$} & {PSNR $\uparrow$} & {SSIM $\uparrow$} & {MAE $\downarrow$} \\
\midrule
Gamma-World (Bidirectional) & POV & 333.4 & 67.25 & 0.557 & 12.87 & 0.453 & 37.70 \\
Gamma-World (Causal) & POV & 323.7 & 70.16 & 0.572 & 13.16 & 0.435 & 37.67 \\
Gamma-World (Distilled) & POV & 340.1 & 69.14 & 0.551 & 12.98 & 0.416 & 38.12 \\
Solaris & POV & 482.7 & 79.50 & 0.593 & 11.49 & 0.429 & 46.33 \\
MultiWorld & POV & 715.8 & 185.33 & 0.642 & 12.68 & 0.381 & 43.49 \\
PAWA (Causal) & POV & \bfseries 157.0 & \bfseries 41.61 & \bfseries 0.380 & \bfseries 16.95 & \bfseries 0.546 & \bfseries 20.97 \\
PAWA (Causal) & Panorama & \bfseries 158.8 & \bfseries 35.91 & \bfseries 0.209 & \bfseries 20.75 & \bfseries 0.653 & \bfseries 12.03 \\
PAWA (Distilled) & POV & \snd{000.0}{175.0} & \snd{000.00}{46.18} & \snd{0.000}{0.437} & \snd{00.00}{15.80} & \snd{0.000}{0.502} & \snd{00.00}{26.62} \\
PAWA (Distilled) & Panorama & \snd{000.0}{166.5} & \snd{000.00}{37.72} & \snd{0.000}{0.226} & \snd{00.00}{20.08} & \snd{0.000}{0.623} & \snd{00.00}{13.30} \\
\bottomrule\end{tabular}
\end{table}

\begin{table}[t]
\caption{Four-player tier with 12 episodes and 239 forecast frames per player. \textbf{Bold} and \underline{underline} mark the best and second best within each stream and evaluation group.}
\label{tab:four-player}
\centering\footnotesize\setlength{\tabcolsep}{3.5pt}
\begin{tabular}{@{}llS[table-format=3.1]S[table-format=3.2]S[table-format=1.3]S[table-format=2.2]S[table-format=1.3]S[table-format=2.2]@{}}
\toprule
Model & Stream & {FVD $\downarrow$} & {FID $\downarrow$} & {LPIPS $\downarrow$} & {PSNR $\uparrow$} & {SSIM $\uparrow$} & {MAE $\downarrow$} \\
\midrule
Gamma-World (Bidirectional) & POV & 752.3 & 142.95 & 0.633 & 13.34 & 0.487 & 38.14 \\
Gamma-World (Causal) & POV & 373.0 & 57.02 & 0.587 & 13.06 & 0.465 & 40.38 \\
Gamma-World (Distilled) & POV & 323.6 & 56.65 & 0.577 & 12.39 & 0.440 & 42.73 \\
PAWA (Causal) & POV & \bfseries 138.0 & \bfseries 33.49 & \bfseries 0.427 & \bfseries 16.41 & \bfseries 0.550 & \bfseries 24.68 \\
PAWA (Causal) & Panorama & \bfseries 161.9 & \bfseries 32.79 & \bfseries 0.254 & \bfseries 19.08 & \bfseries 0.630 & \bfseries 17.42 \\
PAWA (Distilled) & POV & \snd{000.0}{167.3} & \snd{000.00}{33.86} & \snd{0.000}{0.481} & \snd{00.00}{15.08} & \snd{0.000}{0.506} & \snd{00.00}{30.54} \\
PAWA (Distilled) & Panorama & \snd{000.0}{194.1} & \snd{000.00}{32.94} & \snd{0.000}{0.264} & \snd{00.00}{18.67} & \snd{0.000}{0.619} & \snd{00.00}{17.79} \\
\bottomrule\end{tabular}
\end{table}

\begin{table}[t]
\caption{Eight- and sixteen-player tiers with 12 episodes each and 239 forecast frames per player. At eight players, Gamma-World runs with its simplex player encoding extended from four to eight codes. \textbf{Bold} and \underline{underline} mark the best and second best within each stream and evaluation group.}
\label{tab:population}\centering\footnotesize\setlength{\tabcolsep}{3.5pt}
\begin{tabular}{@{}llS[table-format=4.1]S[table-format=3.2]S[table-format=1.3]S[table-format=2.2]S[table-format=1.3]S[table-format=2.2]@{}}\toprule
Model & Stream & {FVD $\downarrow$} & {FID $\downarrow$} & {LPIPS $\downarrow$} & {PSNR $\uparrow$} & {SSIM $\uparrow$} & {MAE $\downarrow$} \\\midrule
\multicolumn{8}{l}{\textit{8 players}} \\
Gamma-World (Bidirectional) & POV & 1707.7 & 213.93 & 0.681 & 12.83 & \snd{0.000}{0.497} & 43.22 \\
Gamma-World (Causal) & POV & 297.3 & 48.84 & 0.606 & 12.21 & 0.453 & 45.16 \\
Gamma-World (Distilled) & POV & 267.1 & 48.27 & 0.586 & 11.56 & 0.432 & 47.28 \\
PAWA (Causal) & POV & \bfseries 111.5 & \snd{000.00}{32.32} & \bfseries 0.458 & \bfseries 15.06 & \bfseries 0.536 & \bfseries 28.92 \\
PAWA (Causal) & Panorama & \bfseries 127.9 & \bfseries 27.42 & \bfseries 0.294 & \bfseries 17.31 & \bfseries 0.591 & \bfseries 20.52 \\
PAWA (Distilled) & POV & \snd{0000.0}{141.3} & \bfseries 28.90 & \snd{0.000}{0.517} & \snd{00.00}{13.71} & 0.494 & \snd{00.00}{35.51} \\
PAWA (Distilled) & Panorama & \snd{0000.0}{135.6} & \snd{000.00}{29.52} & \snd{0.000}{0.301} & \snd{00.00}{16.72} & \snd{0.000}{0.577} & \snd{00.00}{22.57} \\
\midrule
\multicolumn{8}{l}{\textit{16 players}} \\
PAWA (Causal) & POV & \bfseries 92.2 & \snd{000.00}{28.39} & \bfseries 0.467 & \bfseries 15.50 & \bfseries 0.554 & \bfseries 28.49 \\
PAWA (Causal) & Panorama & \snd{0000.0}{93.6} & \snd{000.00}{23.65} & \bfseries 0.333 & \bfseries 16.87 & \bfseries 0.570 & \bfseries 22.58 \\
PAWA (Distilled) & POV & \snd{0000.0}{116.7} & \bfseries 23.96 & \snd{0.000}{0.520} & \snd{00.00}{14.15} & \snd{0.000}{0.511} & \snd{00.00}{35.04} \\
PAWA (Distilled) & Panorama & \bfseries 92.5 & \bfseries 20.95 & \bfseries 0.333 & \snd{00.00}{16.66} & \snd{0.000}{0.567} & \snd{00.00}{22.79} \\
\bottomrule\end{tabular}\end{table}

\subsection{Out-of-view changes, action following and player codes}
\label{app:out-of-view}

\textbf{Co-players entering the view.} Table~\ref{tab:reentry} isolates the situation that motivates
panoramic prediction. From the recorded poses, we find events in which a co-player stays outside
the frustum of a player for at least one second and then enters it, and we score the co-player
region of the POV over the next ten frames with the masks of Appendix~\ref{app:drift}. Across 48
events from both two-player tiers and 111 four-player and 505 eight-player events, \modelname{}
(Distilled) reconstructs the entering co-player with lower LPIPS than every baseline and than the static copy of the
conditioning frame, and the 95\% intervals of the paired difference from the best baseline exclude zero
at every population. The advantage holds when the viewer turns at least $45^\circ$ or the
co-player moves at least 2\,m while hidden, and for regions revealed by camera rotation for the
first time, where \modelname{} improves LPIPS over the best baseline by 0.13, 0.10 and 0.09 at
two, four and eight players.

\begin{table}[t]
\centering
\caption{Co-player LPIPS in the POV over the 10 frames after a co-player enters a player's view
after at least 20 frames outside it. Static repeats the conditioning frame, and
the last column is the paired change of \modelname{} (Distilled) against the best baseline.}
\label{tab:reentry}
\vspace{4pt}\footnotesize\setlength{\tabcolsep}{4pt}
\begin{tabular}{@{}S[table-format=1]S[table-format=3]S[table-format=1.3]S[table-format=1.3]S[table-format=1.3]S[table-format=+1.3]@{\,}l@{}}
\toprule
{Players} & {Events} & {\modelname{}} & {Best baseline} & {Static} & \multicolumn{2}{c}{\modelname{} $-$ best} \\
\midrule
2 & 48 & \bfseries 0.582 & 0.654 & 0.635 & -0.071 & $[-0.116,-0.023]$ \\
4 & 111 & \bfseries 0.603 & 0.649 & 0.647 & -0.046 & $[-0.076,-0.018]$ \\
8 & 505 & \bfseries 0.649 & 0.710 & 0.688 & -0.061 & $[-0.081,-0.040]$ \\
\bottomrule
\end{tabular}
\end{table}

\textbf{Co-players in the panorama.} The panorama of \modelname{} covers every direction around
the player, and its co-player regions reach a similar LPIPS whether or not the co-player lies inside the POV
frustum of the player, with a difference of $+0.009$ $[-0.013,0.031]$ on the interaction tier.
For co-players outside the POV, the panorama improves on the static first panorama by 0.086,
0.031 and 0.038 at two, four and eight players. Changes outside the local view are thus
forecast in the panorama.

\textbf{Action following and binding.} Table~\ref{tab:action-following} estimates the camera
rotation of each generated POV by phase correlation between consecutive frames with HUD rows
removed, sums it over 0.5\,s windows and compares it with the same estimate on the recorded
video. This estimate correlates with the commanded yaw at 0.80 to 0.91. \modelname{} follows the camera
actions with the lowest error of all evaluated models at every population, reducing the error of
the best baseline by 33 to 43\% for the causal model and by 14 to 31\% for the distilled model.
For action binding, a player counts as bound when its own actions explain the rotation of its
stream better than the actions of any co-player whose camera turns noticeably.
\modelname{} binds 90 to 98\% of the players in populations of up to eight. When two players swap
action streams while keeping their initial observations and player codes, 95\% of the swapped
two-player streams are best explained by the injected actions, and the generated POVs thus turn
with the given actions.

\begin{table}[!htb]
\centering
\caption{Action following. Top, mean absolute error in degrees per 0.5\,s window between the yaw
rotation estimated from the generated POV and the same estimate on the recorded video. Bottom,
the fraction of players whose own actions best explain the rotation of their generated stream
among all players whose camera turns noticeably, with chance in parentheses and the recorded videos as a reference.}
\label{tab:action-following}
\vspace{4pt}\footnotesize\setlength{\tabcolsep}{4pt}
\begin{tabular}{@{}lccS[table-format=2.2]S[table-format=2.2]S[table-format=2.2]@{}}
\toprule
& \multicolumn{5}{c}{Players} \\
\cmidrule(l){2-6}
Yaw error $\downarrow$ & 2, low-motion & 2, interaction & 4 & 8 & 16 \\
\midrule
Gamma-World (Bidirectional) & 4.65 & 8.55 & 11.16 & 12.23 & {N/A} \\
Gamma-World (Causal) & 4.77 & 8.80 & 8.55 & 8.45 & {N/A} \\
Gamma-World (Distilled) & 3.68 & 8.35 & 7.77 & 8.00 & {N/A} \\
Solaris & 7.72 & 13.10 & {N/A} & {N/A} & {N/A} \\
MultiWorld & 6.26 & 13.90 & {N/A} & {N/A} & {N/A} \\
\modelname{} (Causal) & \textbf{2.10} & \textbf{4.81} & \bfseries 4.83 & \bfseries 5.32 & \bfseries 6.75 \\
\modelname{} (Distilled) & \underline{3.17} & \underline{5.79} & \snd{00.00}{5.38} & \snd{00.00}{6.74} & \snd{00.00}{7.75} \\
\midrule
Action binding $\uparrow$ & & 2 (0.50) & {4 (0.26)} & {8 (0.13)} & {16 (0.06)} \\
\midrule
Recorded videos & & 0.98 & 0.98 & 0.96 & 0.86 \\
\modelname{} (Causal) & & 0.95 & 0.98 & 0.92 & 0.83 \\
\modelname{} (Distilled) & & 0.93 & 0.94 & 0.90 & 0.73 \\
\bottomrule
\end{tabular}
\end{table}

\textbf{Player code families.}
\label{app:code-assignment}
The codebook reaches $m^2-1$ codes by combining code families whose mutual correlation is
$m^{-1/2}$. Resampling \modelname{} (Distilled) with codes drawn from several families instead
of one, with the same checkpoint, noise and sampler, changes POV FVD by $-6.8$,
$-11.6$ and $-17.0$ at two, four and eight players, and the player pool thus extends beyond a
single family.

\begin{figure}[t]\centering
\includegraphics[width=0.95\linewidth]{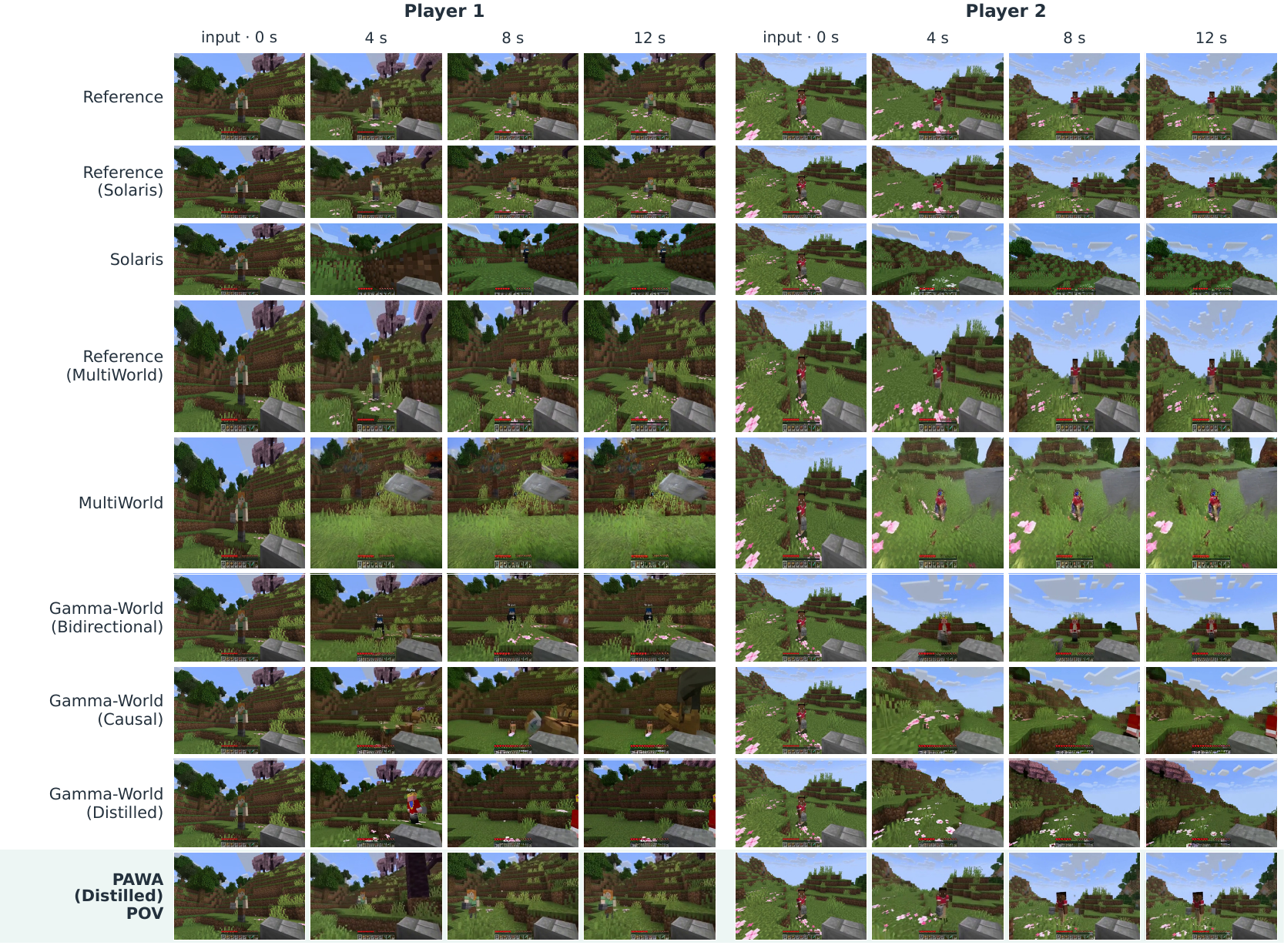}
\caption{\textbf{Two-player comparison with all baselines.} The episode of
Figure~\ref{fig:rollout} with frames 0, 80, 160 and 239 at 20\,Hz. Separate Solaris and
MultiWorld references preserve their native preprocessing. Gamma-World (Bidirectional) and
Gamma-World (Causal) use 35 steps, MultiWorld uses 50 denoising steps per block and the other models use four.}
\label{fig:rollout-external-showcase}\end{figure}

\begin{figure}[t]\centering
\includegraphics[width=0.95\linewidth]{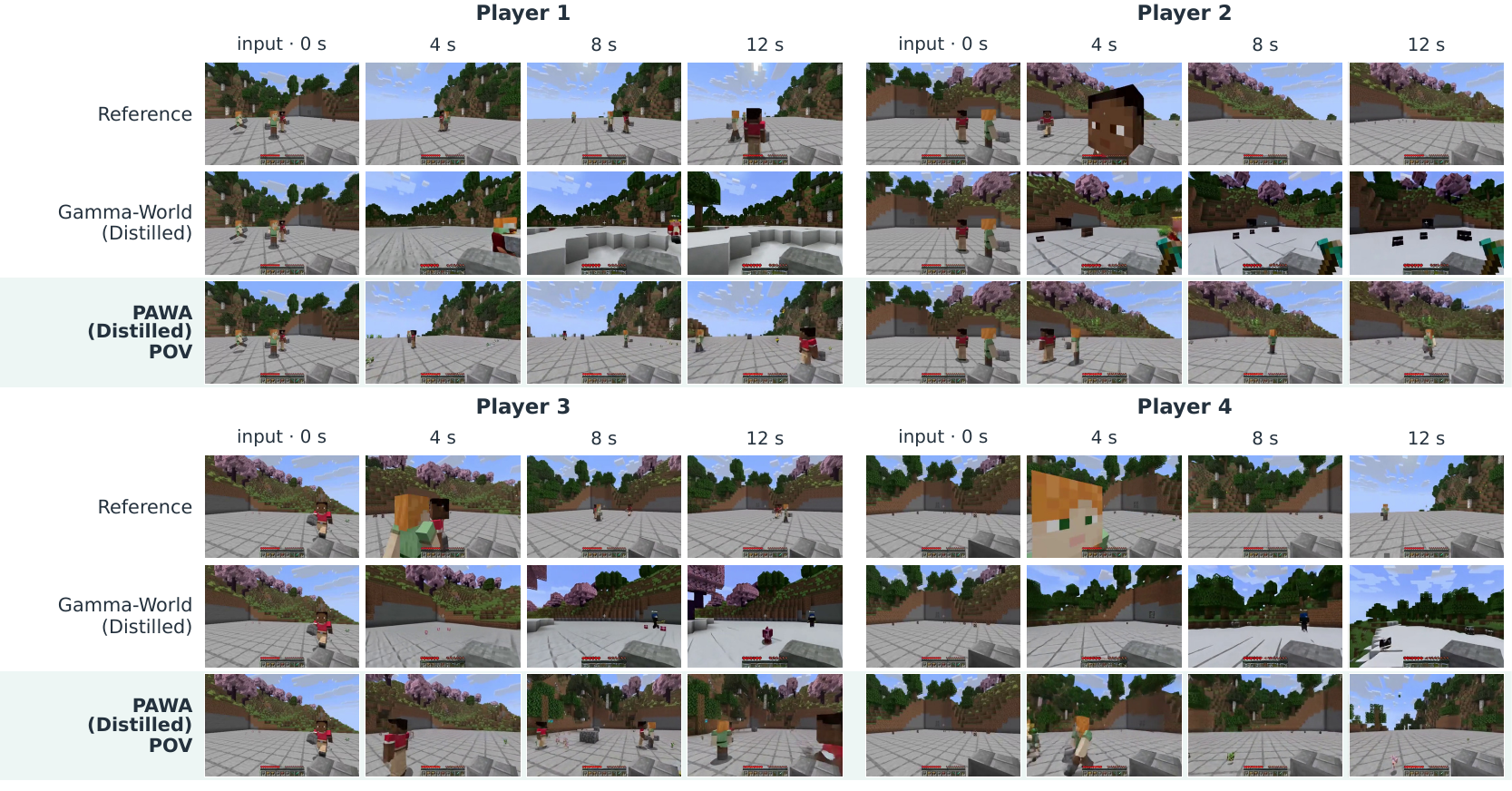}
\caption{\textbf{Synchronized four-player forecasts.} Both distilled models use four steps, and
the first column contains the recorded inputs of the four players.}
\label{fig:rollout-four-player}\end{figure}

\subsection{Ablation of the ray-aligned cross-attention}
\label{app:interventions}

\textbf{Removing the cross-attention.} Table~\ref{tab:inference-sites} disables all seven
cross-attention sites of the final distilled checkpoint without retraining, with identical
episodes, conditioning inputs, sampler, seed and recorded noise. With the sites enabled, POV FVD is
23\% lower, and the paired POV LPIPS change is $-0.045$ with a 95\% episode-bootstrap interval of
$[-0.058,-0.032]$. On the 24 episodes outside checkpoint selection, the change is $-0.048$
$[-0.063,-0.033]$, with improvements on 22 episodes.

\begin{table}[t]
\centering
\small
\caption{Inference-time removal of the seven cross-attention sites from PAWA (Distilled). Both configurations use the same final distilled checkpoint, 30 two-player interaction episodes, recorded noise, seed 1 and four-step sampler. No model is retrained. \textbf{Bold} and \underline{underline} mark the best and second best within each stream and evaluation group.}
\label{tab:inference-sites}
\begin{tabular}{llS[table-format=3.1]S[table-format=2.2]S[table-format=1.4]S[table-format=2.3]S[table-format=1.4]}
\toprule
Stream & Sites & {FVD$\downarrow$} & {FID$\downarrow$} & {LPIPS$\downarrow$} & {PSNR$\uparrow$} & {SSIM$\uparrow$}\\
\midrule
POV & Enabled & \bfseries 138.1 & \bfseries 28.40 & \bfseries 0.4648 & \bfseries 15.382 & \bfseries 0.5157 \\
POV & Disabled & \snd{000.0}{179.0} & \snd{00.00}{37.71} & \snd{0.0000}{0.5096} & \snd{00.000}{14.749} & \snd{0.0000}{0.5094} \\
Panorama & Enabled & \bfseries 98.1 & \bfseries 28.26 & \bfseries 0.2425 & \bfseries 19.765 & \bfseries 0.6172 \\
Panorama & Disabled & \snd{000.0}{104.0} & \snd{00.00}{28.42} & \snd{0.0000}{0.2444} & \snd{00.000}{19.645} & \snd{0.0000}{0.6142} \\
\bottomrule
\end{tabular}
\end{table}

\textbf{Directions and sites.}
\label{app:exchange-directions}
Table~\ref{tab:exchange-directions} removes each cross-attention direction and each group of
sites separately. Removing only the pano$\to$POV direction reproduces the POV degradation of
removing every site, since the POV stream then receives no panoramic features, whereas
removing only the POV$\to$pano direction raises panoramic FVD by 8.0, the same as removing every
site. The early and late sites contribute comparably, and removing all sites degrades POV LPIPS on
29 of 30 episodes.

\begin{table}[t]
\centering
\caption{Inference-time removal of cross-attention directions and sites from \modelname{}
(Distilled) on the 30 two-player interaction episodes. Values are changes against the outputs of the full model
regenerated in the same run with identical checkpoint, recorded noise and sampler, and positive
FVD and LPIPS changes mean degradation. Pano$\to$POV denotes POV queries reading panoramic keys,
early sites follow blocks 3, 7 and 11, and late sites follow blocks 19, 23 and 27.}
\label{tab:exchange-directions}
\vspace{4pt}\footnotesize\setlength{\tabcolsep}{3.5pt}
\begin{tabular}{@{}lS[table-format=+2.1]S[table-format=+2.2]S[table-format=+1.3]cS[table-format=+1.2]S[table-format=+1.1]S[table-format=+1.3]@{}}
\toprule
& \multicolumn{5}{c}{POV} & \multicolumn{2}{c}{Panorama} \\
\cmidrule(lr){2-6}\cmidrule(l){7-8}
Removed & {$\Delta$FVD} & {$\Delta$FID} & {$\Delta$LPIPS} & 95\% interval & {$\Delta$PSNR} & {$\Delta$FVD} & {$\Delta$LPIPS} \\
\midrule
All sites & +36.0 & +10.04 & +0.052 & $[0.039,0.067]$ & -1.09 & +8.0 & +0.002 \\
Pano$\to$POV only & +36.0 & +10.04 & +0.052 & $[0.039,0.067]$ & -1.09 & +1.2 & -0.001 \\
POV$\to$pano only & +3.8 & +0.86 & +0.009 & $[0.004,0.014]$ & -0.34 & +8.0 & +0.002 \\
Early sites & +10.3 & +3.87 & +0.024 & $[0.013,0.035]$ & -0.58 & -2.9 & -0.001 \\
Late sites & +6.8 & +2.61 & +0.025 & $[0.016,0.035]$ & -0.56 & +6.0 & +0.001 \\
\bottomrule
\end{tabular}
\end{table}

\textbf{Training without the cross-attention.}
\label{app:train-ablation}
The removal above measures how much the trained model relies on the cross-attention. To test
whether training with it improves prediction, we continue the causal model from the iteration-3000
checkpoint of its training run in two arms that share the code, learning rate, batch and data
distribution and differ in whether the cross-attention sites are present
(Table~\ref{tab:train-ablation}). The arms see identical batches for the first 2000 updates, after
which the arm without sites resumes with a new batch order following an evaluation pause. Both are
sampled with the causal 35-step sampler on the 30 two-player interaction episodes. Disabling the
sites of the starting checkpoint raises POV LPIPS by 0.012, and training without them does not
close this gap but widens it, to $0.030$ $[0.016,0.043]$ after 2000 updates and $0.056$
$[0.039,0.073]$ after 3000 updates. At that point the arm with cross-attention also leads by 53 in
FVD and $1.9$\,dB in PSNR, with lower LPIPS in 26 of the 30 episodes. The arm without
cross-attention increasingly collapses late in the rollout into over-scaled avatars and uniform
views, while the arm with it keeps the scene layout of the reference. Training with the
ray-aligned cross-attention thus improves POV prediction in LPIPS, FID and PSNR, and the gap after
training exceeds the gap caused by removing the sites alone.

\begin{table}[t]
\centering
\small
\caption{Training with and without the cross-attention. Both arms continue the causal model from
the same checkpoint with the same data distribution, learning rate and batch size, and differ in the
cross-attention sites. Their batches are identical for the first 2000 updates. At 0 updates the arm without cross-attention is the starting checkpoint
with its sites disabled. POV scores on the 30 two-player interaction episodes. \textbf{Bold} marks
the better arm at each training length.}
\label{tab:train-ablation}
\begin{tabular}{S[table-format=4]cS[table-format=3.1]S[table-format=2.2]S[table-format=1.3]S[table-format=2.2]}
\toprule
{Updates} & Cross-attention & {FVD$\downarrow$} & {FID$\downarrow$} & {LPIPS$\downarrow$} & {PSNR$\uparrow$}\\
\midrule
0 & with & \bfseries 103.5 & \bfseries 27.68 & \bfseries 0.452 & \bfseries 16.08 \\
0 & without & 131.1 & 30.67 & 0.464 & 15.90 \\
\midrule
1000 & with & \bfseries 139.8 & \bfseries 29.98 & \bfseries 0.455 & \bfseries 15.90 \\
1000 & without & 152.1 & 34.23 & 0.462 & 15.66 \\
\midrule
2000 & with & \bfseries 143.7 & \bfseries 32.35 & \bfseries 0.448 & \bfseries 16.14 \\
2000 & without & 164.7 & 38.22 & 0.478 & 15.56 \\
\midrule
3000 & with & \bfseries 138.4 & \bfseries 34.68 & \bfseries 0.455 & \bfseries 16.01 \\
3000 & without & 191.7 & 40.69 & 0.511 & 14.08 \\
\bottomrule
\end{tabular}
\end{table}

\textbf{Learned ray alignment.}
\label{app:ray-alignment}
Table~\ref{tab:ray-alignment} measures where POV queries attend on the panorama during sampling on
eight interaction episodes. Under uniform attention, the expected angle between a query ray and
the attended key rays is $90^\circ$ and 10\% of the mass falls inside the POV frustum. At sites 19
and 23 of the distilled model, 43 to 46\% of the mass falls inside the frustum and the most
attended key lies a median of $7^\circ$ to $9^\circ$ from the query ray, and the causal model
reaches $5^\circ$. At the late sites, the reverse direction is aligned as well, with in-frustum panoramic queries
attending to POV keys at an expected angle of $21^\circ$ to $36^\circ$ against a chance level of
$44^\circ$. Re-initializing the cross-attention sites while keeping the ray encoding returns every
site to chance, and the directional alignment emerges during training without an auxiliary alignment loss.

\begin{table}[!htb]
\centering
\caption{Geometry of pano$\to$POV attention at the late cross-attention sites, averaged over
heads. For each POV query we report the expected angle between its ray and the attended panoramic
key rays, the attention mass inside the POV frustum, the mass within $15^\circ$ of the query ray,
and the median angle to the most attended key. Chance is uniform attention over the panorama, and
the control re-initializes the cross-attention sites of the distilled model while keeping its ray
encoding.}
\label{tab:ray-alignment}
\vspace{4pt}\footnotesize\setlength{\tabcolsep}{4pt}
\begin{tabular}{@{}llrS[table-format=1.2]S[table-format=1.3]r@{}}
\toprule
Model & Site & Angle $\downarrow$ & {Frustum mass $\uparrow$} & {Mass within $15^\circ$ $\uparrow$} & Argmax error $\downarrow$ \\
\midrule
\modelname{} (Distilled) & 19 & $51^\circ$ & 0.43 & 0.25 & $9^\circ$ \\
& 23 & $50^\circ$ & 0.46 & 0.29 & $7^\circ$ \\
& 27 & $65^\circ$ & 0.28 & 0.16 & $24^\circ$ \\
\midrule
\modelname{} (Causal) & 19 & $47^\circ$ & 0.47 & 0.29 & $8^\circ$ \\
& 23 & $38^\circ$ & 0.57 & 0.43 & $5^\circ$ \\
& 27 & $55^\circ$ & 0.36 & 0.25 & $12^\circ$ \\
\midrule
Control, re-initialized sites & all & $90^\circ$ & 0.10 & 0.011 & ${\approx}90^\circ$ \\
Chance & all & $90^\circ$ & 0.10 & 0.012 & ${\approx}90^\circ$ \\
\bottomrule
\end{tabular}
\end{table}

\subsection{Qualitative comparisons}
Figure~\ref{fig:rollout-external-showcase} shows the two-player episode of
Figure~\ref{fig:rollout} with every baseline, and Figure~\ref{fig:rollout-four-player} shows a
four-player episode. Both episodes were chosen for a recognizable scene with co-players visible in
the generated POVs, and all videos are the evaluated outputs. Player groups share input and
forecast columns as in \citet{liu2026gammaworld}, and Reference denotes the recorded continuation.

\subsection{Co-player fidelity over long rollouts}
\label{app:drift}
\begin{table}[!htb]
\centering
\caption{POV LPIPS by forecast-frame band on the 30 two-player interaction episodes, over the
whole frame and within projected co-player regions. Repeat frame 0 copies the conditioning frame
for the whole rollout and serves as a static reference.}
\label{tab:coplayer-bands}
\vspace{4pt}\footnotesize\setlength{\tabcolsep}{3.5pt}
\begin{tabular}{@{}lS[table-format=1.3]S[table-format=1.3]S[table-format=1.3]S[table-format=1.3]S[table-format=1.3]S[table-format=1.3]S[table-format=1.3]S[table-format=1.3]@{}}
\toprule
& \multicolumn{4}{c}{Whole frame} & \multicolumn{4}{c}{Co-player regions} \\
\cmidrule(lr){2-5}\cmidrule(l){6-9}
Model & {1--44} & {45--92} & {93--188} & {189--239} & {1--44} & {45--92} & {93--188} & {189--239} \\
\midrule
Gamma-World (Bidirectional) & 0.447 & 0.565 & 0.595 & 0.621 & 0.505 & 0.617 & 0.656 & 0.646 \\
Gamma-World (Causal) & 0.452 & 0.591 & 0.645 & 0.678 & 0.514 & 0.609 & 0.679 & 0.710 \\
Gamma-World (Distilled) & 0.446 & 0.571 & 0.617 & 0.651 & 0.507 & 0.607 & 0.661 & 0.718 \\
Solaris & 0.471 & 0.619 & 0.679 & 0.702 & 0.512 & 0.680 & 0.763 & 0.783 \\
MultiWorld & 0.588 & 0.677 & 0.690 & 0.683 & 0.618 & 0.704 & 0.728 & 0.694 \\
PAWA (Causal) & 0.292 & 0.398 & 0.456 & 0.485 & 0.373 & 0.495 & 0.562 & 0.574 \\
PAWA (Distilled) & 0.322 & 0.449 & 0.502 & 0.534 & 0.418 & 0.549 & 0.589 & 0.596 \\
\midrule
Repeat frame 0 & 0.418 & 0.529 & 0.573 & 0.581 & 0.506 & 0.588 & 0.624 & 0.615 \\
\bottomrule
\end{tabular}
\end{table}

Table~\ref{tab:coplayer-bands} reports POV LPIPS in four forecast-frame bands, over the whole
frame and within co-player regions. Co-player regions are projected from the recorded player poses
with the box expansion and temporal union of Appendix~\ref{app:foreground}, HUD rows are excluded,
and a frame counts when the region covers at least 0.5\% of the image. Over the 12-second
rollout, co-player LPIPS rises for every model, by 0.14 to 0.27 for Gamma-World, Solaris and
\modelname{}, and part of every increase reflects the recorded future diverging from the first
frame, since the LPIPS of the static copy rises by 0.11. In the last band, the co-player
regions of all Gamma-World variants, Solaris and MultiWorld are no closer to the reference than
this static copy, whereas both \modelname{} variants obtain a lower LPIPS. The same trend holds with
four players, where co-player LPIPS rises from 0.476 to 0.618 for \modelname{} (Distilled) and
from 0.547 to 0.719 for Gamma-World (Distilled). Loss of co-player fidelity over long rollouts is
therefore common to current multiplayer world models, and \modelname{} keeps the co-player
regions closest to the reference in every band.

\end{document}